\documentclass{article}

    \PassOptionsToPackage{round}{natbib}

    \usepackage[preprint]{neurips_2025}

\usepackage[table]{xcolor}
\usepackage[utf8]{inputenc} 
\usepackage[T1]{fontenc}    
\usepackage{hyperref}       
\usepackage{url}            
\usepackage{booktabs}       
\usepackage{amsfonts}       
\usepackage{nicefrac}       
\usepackage{microtype}      
\usepackage{amsmath}
\usepackage{amssymb}
\usepackage{graphicx}
\usepackage{subfigure}
\usepackage{algorithm}
\usepackage{algorithmic}
\usepackage{multirow}
\usepackage[skins]{tcolorbox}
\usepackage{wrapfig}

\usepackage{enumitem}

\hypersetup{
	colorlinks=true,
	linkcolor=red,
	filecolor=blue,      
	citecolor=teal,
}

\title{Observe-R1: Unlocking Reasoning Abilities \\of MLLMs with Dynamic Progressive \\Reinforcement Learning}

\author{%
  Zirun Guo \\
  Zhejiang University\\
  \small\texttt{zrguo.cs@gmail.com}
  \And
  Minjie Hong \\
  Zhejiang University \\
  \small\texttt{hongminjie@zju.edu.cn} 
  \And
  Tao Jin \\
  Zhejiang University \\
  \small\texttt{jint\_zju@zju.edu.cn} 
}

\begin{document}
\maketitle

\begin{abstract}
Reinforcement Learning (RL) has shown promise in improving the reasoning abilities of Large Language Models (LLMs). However, the specific challenges of adapting RL to multimodal data and formats remain relatively unexplored. In this work, we present Observe-R1, a novel framework aimed at enhancing the reasoning capabilities of multimodal large language models (MLLMs). We draw inspirations from human learning progression—from simple to complex and easy to difficult, and propose a gradual learning paradigm for MLLMs. To this end, we construct the \textit{NeuraLadder} dataset, which is organized and sampled according to the difficulty and complexity of data samples for RL training. To tackle multimodal tasks, we introduce a multimodal format constraint that encourages careful observation of images, resulting in enhanced visual abilities and clearer and more structured responses. Additionally, we implement a bonus reward system that favors concise, correct answers within a length constraint, alongside a dynamic weighting mechanism that prioritizes uncertain and medium-difficulty problems, ensuring that more informative samples have a greater impact on training. Our experiments with the Qwen2.5-VL-3B and Qwen2.5-VL-7B models on 20k samples from the NeuraLadder dataset show that Observe-R1 outperforms a series of larger reasoning models on both reasoning and general benchmarks, achieving superior clarity and conciseness in reasoning chains. Ablation studies validate the effectiveness of our strategies, highlighting the robustness and generalization of our approach. The dataset and code will be released at \url{https://github.com/zrguo/Observe-R1}.

\end{abstract}

\section{Introduction}

\begin{quote}
    \textit{He who would learn to fly one day must first learn to stand and walk and run and climb and dance; one cannot fly into flying.}\hfill\textit{-- Friedrich Nietzsche}
\end{quote}

Reinforcement Learning (RL) has demonstrated great success in eliciting the reasoning abilities of Large Language Models (LLMs)~\citep{shao2024deepseekmath, guo2025deepseek, hu2025reinforce++, xie2025logic, yu2025dapo}. These models such as OpenAI's o1~\citep{jaech2024openai} and DeepSeek-R1~\citep{guo2025deepseek} obtain excellent performance in complex reasoning tasks like mathematical and logical problems through detailed step-by-step analyses. With the advancements in improving reasoning abilities of LLMs, how to enhance the reasoning abilities of multimodal large language models (MLLMs)~\citep{openai2024gpt4ocard, chen2025internvl25, bai2025qwen25vl, zhu2025internvl3} becomes increasingly important.

Recent studies~\citep{huang2025vision, meng2025mm, peng2025lmm, zhang2025r1} have pioneered the exploration of RL to enhance the reasoning abilities of MLLMs. These studies contribute various multimodal datasets and demonstrate that RL training can effectively incentivize the reasoning capabilities of MLLMs. Although these works demonstrate that RL can enhance the reasoning abilities of MLLMs, they only focus on dataset construction or training stage designs and do not make specific adaptation for multimodal data or models.

In this work, we propose Observe-R1, which focuses on two key questions related to enhancing the reasoning capabilities of MLLMs: (1) \textit{How do humans learn to reason in complex tasks, such as solving mathematical problems?} (2) \textit{What are the differences in the reasoning processes between unimodal and multimodal tasks?} For the first question, we contemplate why a PhD student or an expert can tackle complex problems. This ability stems from the fact that they do not engage with difficult problems from the outset; instead, they learn progressively, from easy to difficult and from simple to complex. Through a structured educational journey from elementary school to high school and then to university, they gradually build their understanding, making it easier to comprehend very challenging issues when they arise. Drawing inspiration from this, we propose that the learning of large models should also follow this gradual learning paradigm, avoiding the insufficient utilization of difficult problem data during the early stages of model training. To this end, we consider the difficulty and complexity of each question in the dataset and categorize the dataset according to difficulty and complexity levels, and propose the NeuraLadder dataset for RL training. For the second question, compared to pure language models, MLLMs have more sources of information, requiring a clear understanding of the information from different modalities to answer questions more accurately. Therefore, we propose a multimodal format constraint that encourages the model to first observe the images carefully and think step by step before providing the final answer. This strategy results in responses that are more structured, clear, and concise, while also enhancing visual abilities, training efficiency and speed. Additionally, we introduce a bonus reward term that favors the simplest correct answer within a length constraint, enabling the model to reason correctly as well as more clearly and concisely. Furthermore, we implement a dynamic weighting mechanism based on the model's uncertainty regarding the data samples. This mechanism prioritizes uncertain and medium-difficulty problems while filtering out all-correct and all-incorrect samples, resulting in more stable training.

We conduct extensive experiments using Qwen2.5-VL-3B and Qwen2.5-VL-7B as the base models on 20k data samples from the NeuraLadder dataset and obtain Observe-R1-3B and 7B. Observe-R1-3B outperforms a series of 7-11B reasoning models on math and scientific reasoning benchmarks, despite having only 3B parameters. Additionally, Observe-R1 demonstrates clearer and more concise reasoning chains compared to the baseline models. Furthermore, ablation studies are conducted to explore the strategies individually, confirming the effectiveness and generalization ability of our approach.
Our contributions can be summarized as:
\begin{itemize}[leftmargin=*]
    \setlength{\itemsep}{1pt}
    \setlength{\parsep}{0pt}
    \setlength{\parskip}{0pt}
    \setlength{\leftmargin}{0pt}
    \item \textbf{Innovative RL Framework for MLLMs.} We propose multimodal format, bonus reward and a dynamic weighting mechanism to improve the reasoning capabilities of MLLMs. These strategies enable the model to learn to reason correctly, clearly, concisely and progressively. 
    \item \textbf{Stronger Performance and Generalization Ability.} We develop Observe-R1-3B and 7B. Using only 20k data samples for training, Observe-R1-3B outperforms a series of 7-11B general and reasoning MLLMs on various reasoning and general benchmarks.
    \item \textbf{Open-sourced Dataset, Model and Code.} We construct the NeuraLadder dataset, which is organized and sampled based on the difficulty and complexity of the questions. Our dataset and code are publicly available for further development.
\end{itemize}

\section{Related Work}
\subsection{Reasoning in Large Models} 
Researchers have found that enabling large models to think step-by-step can significantly enhance their performance on reasoning tasks. Common approaches include Chain-of-Thought (CoT) prompting methods~\citep{wei2022chain, wang2022self}, plan-based methods~\citep{yao2023tree}, Monte Carlo Tree Search (MCTS) methods~\citep{feng2023alphazero}, and constructing complicated reasoning supervised fine-tuning dataset~\citep{xu2024llava, muennighoff2025s1}. More recently, RL has become a key technique for directly training reasoning capabilities, moving beyond simple prompting or fine-tuning. DeepSeek-R1~\citep{guo2025deepseek} and OpenAI's o1~\citep{jaech2024openai} represent breakthrough models that leverage RL to develop sophisticated reasoning abilities. The core innovation in these approaches is using verifiable feedback signals to train models on complex reasoning tasks such as mathematical problem-solving and programming challenges. In the multimodal field, recent works~\citep{huang2025vision, meng2025mm, peng2025lmm, zhang2025r1} apply RL to MLLMs to incentivize reasoning abilities and get remarkable results. However, these studies have merely demonstrated that reinforcement learning can effectively enhance the reasoning capabilities of multimodal large models; they have not further optimized or explored the specific characteristics of multimodal models and data.

\subsection{Reinforcement Learning for Large Models} 
RL has emerged as a crucial component in the latest generation of large models~\citep{jaech2024openai, guo2025deepseek}, often leading to superior generalization capabilities~\citep{chu2025sft} compared to purely supervised methods. 
While various RL algorithms exist, the evolution from Trust Region Policy Optimization (TRPO)~\citep{schulman2015trust} to Proximal Policy Optimization (PPO)~\citep{schulman2017proximal}, and more recently to methods like Group Relative Policy Optimization (GRPO)~\citep{shao2024deepseekmath}, reflects a search for stability, sample efficiency, and scalability suitable for massive models. Many recent works~\citep{shen2025dast, yu2025dapo, xie2025logic, hu2025reinforce++} apply these RL algorithms specifically to enhance LLM reasoning and demonstrates great success. For example, DAPO~\citep{yu2025dapo} proposes the decoupled clip and dynamic sampling policy optimization algorithm to avoid entropy collapse and improve training efficiency and stability.

\section{Methodology}
\subsection{Preliminaries}
\textbf{Reinforcement Learning.} Reinforcement learning aims to maximize the following objective:
\begin{equation}\label{eq1}
    \mathcal{J}(\theta)=\mathbb{E}_{y\thicksim\pi_\theta}[r(y)-\beta \mathbb{D}_{\mathrm{KL}}(\pi_\theta||\pi_{\theta_{\text{ref}}})]
\end{equation}
where $y$ is the generated answer, $r(\cdot)$ is the reward function, $\mathbb{D}_{\mathrm{KL}}$ is the KL divergence, $\beta$ is the penalty trade-off and $\pi_\theta$ is the policy model and $\pi_{\theta_\text{ref}}$. In GRPO~\citep{shao2024deepseekmath}, it computes the advantage in a group-relative manner. Concretely, for a question-answer pair $(q,a)$, the policy $\pi_\theta$ samples a group of $G$ distinct responses $\{o_i\}_{i=1}^G$ and then the advantage is computed as:
\begin{equation}\label{eq2}
    \hat{A}_{i}=\frac{r_i-\mathrm{mean}(\{r_i\}_{i=1}^G)}{\mathrm{std}(\{r_i\}_{i=1}^G)}
\end{equation}
where $r_i$ is the rule-based verifiable reward~\citep{guo2025deepseek}. Besides, GRPO adopts a clipped objective following PPO~\citep{schulman2017proximal} and Equation~\ref{eq1} can be rewritten as:
\begin{equation}
    \begin{aligned}
        &\mathcal{J}_{\mathrm{GRPO}}(\theta)  =\mathbb{E}_{(q,a)\thicksim\mathcal{D},\{o_i\}_{i=1}^G\thicksim\pi_{\theta_{\mathrm{old}}}(\cdot|q)} \\
         & \left[\frac{1}{G}\sum_{i=1}^{G}\frac{1}{|o_{i}|}\sum_{t=1}^{|o_{i}|}\left(\min\left(r_{i,t}(\theta)\hat{A}_{i,t},\mathrm{clip}\left(r_{i,t}(\theta),1-\varepsilon,1+\varepsilon\right)\hat{A}_{i,t}\right)-\beta \mathbb{D}_{\mathrm{KL}}(\pi_{\theta}||\pi_{\mathrm{ref}})\right)\right]
    \end{aligned}
\end{equation}
where $r_{i,t}(\theta)$ is the importance sampling term and computed as:
\begin{equation}
    r_{i,t}(\theta)=\frac{\pi_\theta(o_{i,t}\mid q,o_{i,<t})}{\pi_{\theta_{\mathrm{old}}}(o_{i,t}\mid q,o_{i,<t})}
\end{equation}
GRPO does not require critic models and can effectively differentiate between high-quality and low-quality outputs.

\textbf{Rule-based Reward Function.} Rule-based reward has been demonstrated effective in tasks such as mathematical and coding problems~\citep{guo2025deepseek}. DeepSeek-R1 uses two types of rewards (\textit{i.e.,} accuracy and format reward) to incentivize the reasoning abilities of LLMs. Format reward enforces a structured response format. The model is required to follow a specific format to receive the format reward. The accuracy reward evaluates the correctness of the generated answer. If the answer matches the ground truth, the corresponding response will receive the reward.

\begin{figure}
  \vskip 0.2in
  \begin{center}
  \subfigure[]{\includegraphics[width=.34\columnwidth]{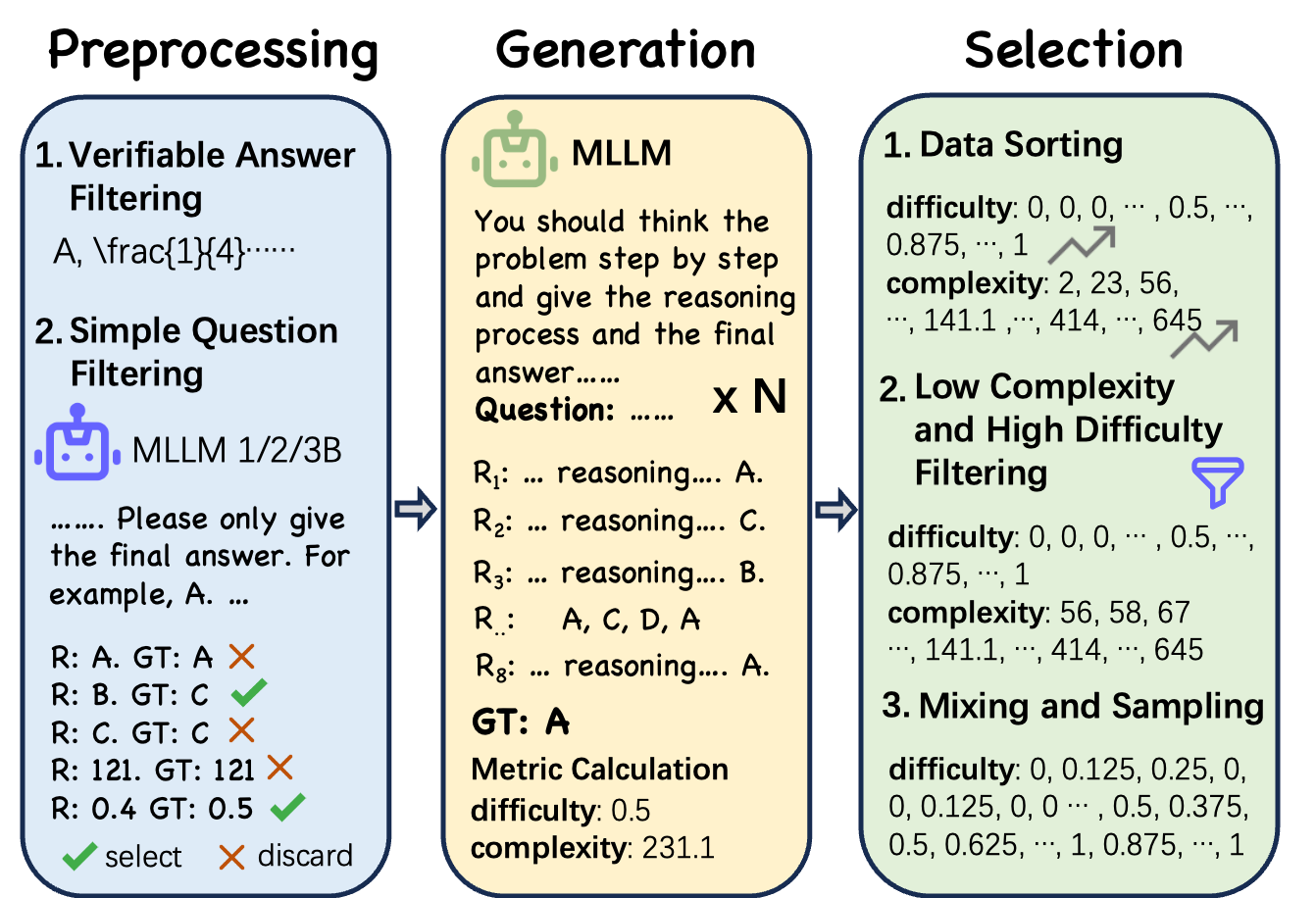}\label{datafig:a}}\hspace{3pt}
 \subfigure[]{\includegraphics[width=.33\columnwidth]{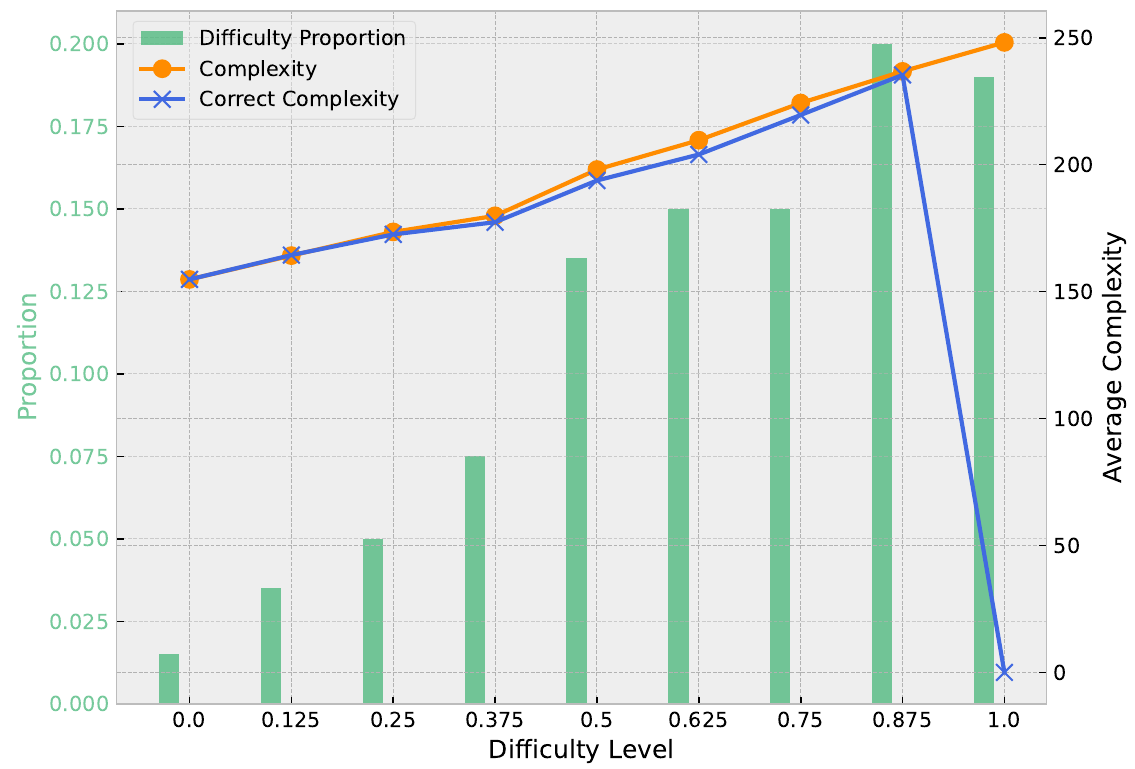}\label{datafig:b}}\hspace{3pt}
 \subfigure[]{\includegraphics[width=.3\columnwidth]{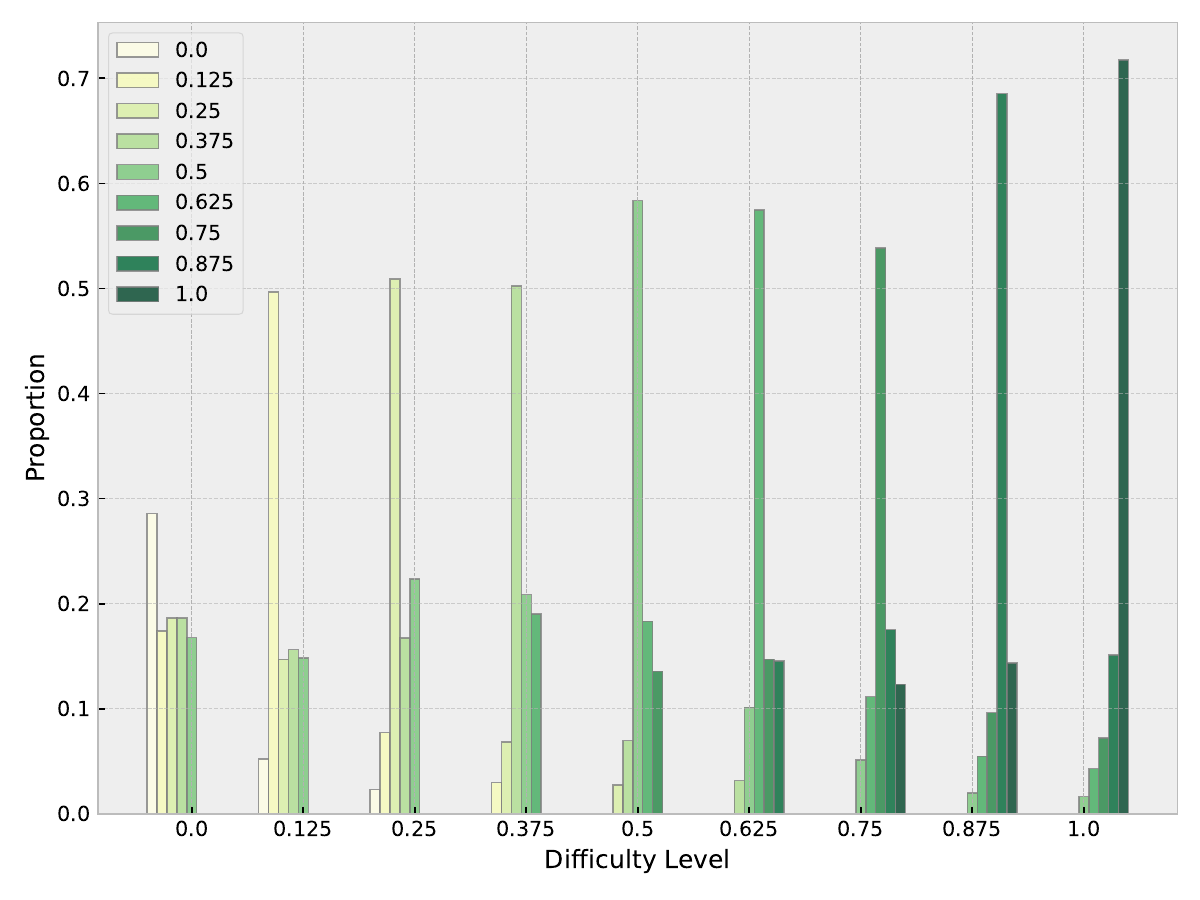}\label{datafig:c}}
\\
  \caption{(a) NeuraLadder construction pipeline. (b) Proportion, average complexity and complexity of the correct responses of the NeuraLadder dataset. (c) Sampled and Mixed results of the dataset.}
  \label{datafig}
  \end{center}
  \vskip -0.15in
\end{figure}

\subsection{NeuraLadder Dataset}
Drawing inspiration from human learning process, we hope that MLLMs can emulate the human learning process by progressing from easy to difficult and from simple to complex, akin to climbing a ladder, gradually increasing the difficulty of questions. Some recent studies utilize difficulty to filter samples for more stable training~\citep{meng2025mm} or to constrain response lengths~\citep{shen2025dast}. For example, a recent study~\citep{shen2025dast} introduces difficulty-adaptive slow-thinking to autonomously adjust the length of Chain-of-Thought (CoT) based on the difficulty of problems. Different from these works, we use the difficulty and complexity to organize and sample the data for progressive learning.
We name our dataset the \textit{NeuraLadder} dataset.
 The construction of the dataset is outlined in Figure~\ref{datafig:a} and can be divided into the following steps: 

\textbf{Data Sources and Preprocessing.} 
We utilize a combination of datasets: MathVision~\citep{wang2024measuring}, We-Math~\citep{qiao2024wemath}, PolyMath~\citep{gupta2024polymath}, SceMQA~\citep{liang2024scemqa} and MathV360K~\citep{shi2024math}. We discard data for which answers cannot be verified. Additionally, we employ Qwen2.5-VL-3B to generate a single answer for each question, providing only the final answer without reasoning and filtering out questions that are answered correctly. By leveraging a relatively small model to generate responses without a detailed reasoning process, we can effectively filter out overly simple questions, thereby reducing the risk of overfitting.

\textbf{Metric Calculation.} For dataset construction, we define two metrics for a question, \textit{difficulty} and \textit{complexity}. Specifically, for a problem in a dataset, we use an MLLM to generate eight responses with reasoning process. Difficulty is calculated as one minus the ratio of correct answers, while complexity is measured as the average length of the eight responses.

\textbf{Data Sorting and Filtering.} After obtaining the dataset with varying levels of difficulty and complexity, we sort the data according to specific criteria: questions with lower difficulty are prioritized, and when difficulty levels are the same, those with lower complexity are selected first.
We select all questions with a difficulty level of 1 (\textit{i.e.,} accuracy is 0) and manually review this subset. We correct any instances where formatting or other errors prevent the retrieval of correct answers, ensuring the quality of the dataset. We do not discard any data with a difficulty level of 1 because we believe that as training progresses, the model will become capable of handling increasingly difficult problems. Therefore, we reserve the filtering of very difficult data for the training process, which will be described in Section~\ref{s333}.
For data with a difficulty level of 0, we discard those with low complexity (\textit{e.g.} complexity lower than 100). After this filtering, the distribution of our data across different difficulty levels is presented in Figure~\ref{datafig:b}.

\textbf{Data Sampling.} To obtain a smoother dataset, we sample and mix data for each difficulty level from its four adjacent difficulty levels, resulting in a smoothed dataset with ordered difficulty. Through this sampling and mixing operation, the model can maintain a preview of higher difficulty data and a review of lower difficulty data during training. Figure~\ref{datafig:c} shows the final mixing ratio for each difficulty level in our dataset.

\subsection{Observe-R1}
\subsubsection{Multimodal Format}
DeepSeek-R1 enforces output format constraints on the model (\textit{i.e.,} \texttt{<think>...</think>} and \texttt{<answer>...</answer>}) using a system prompt. Recent studies~\citep{huang2025vision, meng2025mm} also demonstrate its effectiveness in MLLMs. However, it drives us to think that whether the format for MLLMs should be different for better performance. For a multimodal task such as image-text pairs, a person is expected to observe the image carefully before answering the question. We propose that the clearer the observation, the easier it is to provide accurate answers. To this end, we enforce the MLLM using the following system prompt:

\begin{tcolorbox}[
    colframe=teal!80!black, 
    colback=teal!8!white, 
    coltitle=white, 
    fonttitle=\bfseries,
    title=System Prompt for Multimodal Format, 
    boxrule=0.35mm, 
    drop fuzzy shadow=gray!30 
] 
You are an advanced AI system ... You should first observe the image carefully and think about the reasoning process step by step ... The descriptions of the image required to address the problem, reasoning process and answer are enclosed within <observe> </observe> <think> </think> and <answer> </answer> tags, respectively...
\end{tcolorbox} 

In the format reward, the model is required to strictly follow the above format constraint. With the structured thinking process, we expect the model to produce clearer and more concise responses. Besides, with the \texttt{<observe></observe>} tag, the model is encouraged to observe the image contents carefully, which will in turn improve its visual abilities.

\subsubsection{Additional Bonus Reward}
In Figure~\ref{datafig:b}, we can observe that for data of any difficulty level, the average length of correct answers is consistently lower than the overall average complexity (and also lower than the average complexity of incorrect answers). This phenomenon suggests that excessive reasoning can negatively impact the model's performance, leading to the generation of incorrect answers. Based on this observation, we re-evaluate the reward structure for correct answers. In current common RL methods~\citep{guo2025deepseek, yu2025dapo}, rewards are typically divided into accuracy rewards $r_i^a$ and format rewards $r_i^f$. All correct answers receive the same accuracy reward, thus assigning equal weight to all correct response samples. However, among the correct answers, there are those with clear and concise reasoning chains, as well as those with redundant and excessive reasoning chains. Even among correct samples, overly redundant reasoning chains are detrimental to the model's knowledge reconstruction. Therefore, we propose giving an additional bonus $r_i^b$ to the correct answer with the lowest complexity, encouraging the model's reasoning to converge towards high-quality, clear, and concise thinking. To prevent the generation of incorrectly formatted answers or the absence of a reasoning process, we impose a length constraint on the responses. We only grant this bonus reward when the response length is greater than $\ell$, which can be expressed as:
\begin{equation}\label{eq5}
    \begin{aligned}
    r_i^b=
        \begin{cases}
        1, & \text{if}\ r_i^a=1,\ \ell\leq |o_i|\ \text{and}\  |o_i|=\min(\{|o_j|\}_{j=1}^G) \\
        0, & \text{else} 
        \end{cases}
    \end{aligned}
\end{equation}
where $G$ is the group and $|o_i|$ is the response length. We can flexibly set $\ell$ to achieve different objectives. When $\ell$ is set to a high value, it means that bonus rewards are given only for questions with higher average complexity, while questions with lower complexity do not receive rewards. When $\ell\rightarrow +\infty$, the bonus reward strategy will degenerate into the original reward strategy.
Finally, our total reward can be calculated as:
\begin{equation}
    r_i = r_i^a + \gamma_1r_i^f + \gamma_2r_i^b
\end{equation}
where $\gamma_1$ and $\gamma_2$ are the trade-offs, $r_i^a$ and $r_i^f$ represent the accuracy reward and format reward, respectively. They are assigned a value of 1 when the answer or format is correct, and 0 otherwise. Then advantage will be calculated using Equation~\ref{eq2}.

\subsubsection{Dynamic Weighting and Sampling}\label{s333}
In original GRPO, all samples are assigned equal weight. However, the effort required to solve an easy problem and a medium problem is clearly different. Therefore, we propose assigning different dynamic weights to samples based on their difficulty levels, allowing the model to focus more on optimizing those samples that offer greater benefits during training.

Specifically, we consider one minus the proportion of correct answers generated by the model across $G$ attempts as the difficulty $d$ of the question. For a model, medium-difficulty questions have the greatest uncertainty, because for easy questions, the model can generate correct answers multiple times, while for difficult questions, the model cannot generate correct answers even after multiple attempts. Therefore, both easy and difficult questions have relatively high certainty for the model. During training, we need to ensure that the model reduces uncertainty about medium-difficulty questions and focuses on optimizing samples with low certainty, so we need to increase their weight. Therefore, we propose to grant the greatest weight to samples with a difficulty level of $d=0.5$ and decrease the weight as the difficulty $d$ moves further away from $d=0.5$. At the same time, as shown in Figure~\ref{rrspon:a}, as training progresses, the model's reasoning ability becomes stronger, and it can easily solve those problems that are originally of medium difficulty. Therefore, during training, medium-difficulty problems will gradually become easy ones, and difficult problems will gradually become medium-difficulty ones. This ensures that our strategy not only focuses on medium-difficulty problems but also dynamically and indirectly focuses on those problems that are originally difficult for the base model.

Therefore, we construct a symmetric function about difficulty $d$ with an axis of symmetry at $d=0.5$, such that the weight reaches its maximum value at $f(0.5)$, and decreases as the difficulty d moves further away from $d=0.5$. At the same time, in order to filter out all-correct and all-incorrect samples for better training stability, we set $f(0)=f(1)=0$, so that the function can also act as a mask, filtering out all-correct and all-incorrect samples to make training more stable. Finally, we select our function as $f(d) = 4\sigma d(1-d)$ where $\sigma$ is a coefficient for trade-off. Following previous studies~\citep{meng2025mm, yu2025dapo}, we do not include KL loss in our objective. Therefore, our final optimization objective can be expressed as:
\begin{equation}
    \begin{aligned}
        &\mathcal{J}(\theta)  =\mathbb{E}_{(q,a)\thicksim\mathcal{D},\{o_i\}_{i=1}^G\thicksim\pi_{\theta_{\mathrm{old}}}(\cdot|q)} \\
         & \left[\frac{1}{G}\sum_{i=1}^{G}\frac{{\color{blue} f(d)}}{|o_{i}|}\sum_{t=1}^{|o_{i}|}\left( \min\left(r_{i,t}(\theta){\color{blue}\hat{A}_{i,t}},\mathrm{clip}\left(r_{i,t}(\theta),1-\varepsilon,1+\varepsilon\right){\color{blue}\hat{A}_{i,t}}\right)\right)\right]
    \end{aligned}
\end{equation}
where $\color{blue}\text{blue}$ denotes the differences between GRPO and our algorithm.

\begin{table}[]
\centering
\caption{Comparisons with SOTA MLLMs, including closed-source models, open-source general models and reasoning models.}
\label{mainres}
\resizebox{0.98\columnwidth}{!}{%
\begin{tabular}{@{}l|ccc|ccc@{}}
\toprule
\multicolumn{1}{c}{\multirow{2}{*}{Model}} & \multicolumn{3}{c}{Reasoning   Tasks} & \multicolumn{3}{c}{General Tasks} \\ \cmidrule(l){2-7} 
\multicolumn{1}{c}{} & MathVista & MathVerse & MMK12 & MMStar & MMMU & MMBench \\ \midrule
\multicolumn{7}{l}{\textit{Closed-Source Model}} \\\midrule
GPT-4V~\citep{openai2023GPT4V} & 49.4 & 39.4 & - & 49.7 & - & - \\
GPT-4o~\citep{openai2024gpt4ocard} & 63.8 & 37.6 & 49.9 & 61.6 & - & - \\
Claude-3.5 Sonnet~\citep{anthro2024claude} & 67.7 & 26.5 & - & 62.2 & - & - \\\midrule
\multicolumn{7}{l}{\textit{Open-Source Model}} \\\midrule
InternVL2.5-8B~\citep{chen2025internvl25} & 64.4 & 39.5 & 45.6 & 62.8 & 56.2 & - \\\midrule
\multicolumn{7}{l}{\textit{Reasoning Model}} \\\midrule
R1-Onevision-7B~\citep{yang2025r1} & 64.1 & 46.4 & 39.8 & 61.9 & 44.4 & - \\
R1-VL-7B~\citep{zhang2025r1} & 63.5 & 40.0 & 32.7 & 60.0 & 43.3 & - \\
LLaVA-o1-11B~\citep{xu2024llava} & 54.8 & 20.3 & - & 57.9 & - & - \\
LlamaV-o1-11B~\citep{thawakar2025llamav} & 54.4 & - & - & 59.4 & - & - \\
Math-LLaVA-13B~\citep{shi2024math} & 46.4 & 22.9 & - & - & - & - \\\midrule
Qwen2.5-VL-3B~\citep{bai2025qwen25vl} & 57.8 & 34.9 & 43.5 & 54.1 & 47.6 & 63.7 \\
Qwen2.5-VL-3B-GRPO & 60.0 & 38.2 & 44.2 & 57.9 & 45.4 & 60.2 \\
\rowcolor{gray!30} Observe-R1-3B (Ours) & 64.9 & 40.7 & 45.3 & 60.7 & 48.6 & 64.1  \\\midrule
Qwen2.5-VL-7B~\citep{bai2025qwen25vl} & 64.8 & 46.7 & 53.1 & 64.1 & 48.4 & 71.8 \\
\rowcolor{gray!30} Observe-R1-7B (Ours) & 68.4 & 50.8 & 55.1 & 66.2 & 49.5 & 72.8 \\ \bottomrule
\end{tabular}%
}
\end{table}

\section{Experiments}
\subsection{Implementation Details}
During the data construction stage, we use Qwen2.5-VL-3B to filter out overly simple problems and Qwen2.5-VL-7B to generate the reasoning processes for calculating difficulty and complexity. We use Qwen2.5-VL-7B to generate the reasoning processes because it functions as a medium-level AI assistant. The 3B version may not accurately capture the difficulty of medium and hard problems, while larger models, such as the 72B version, might fail to reflect the difficulty of easier problems. In contrast, the 7B version provides a relatively accurate assessment of data difficulty.

\begin{figure*}
  \begin{center}
  \centerline{\includegraphics[width=0.75\columnwidth]{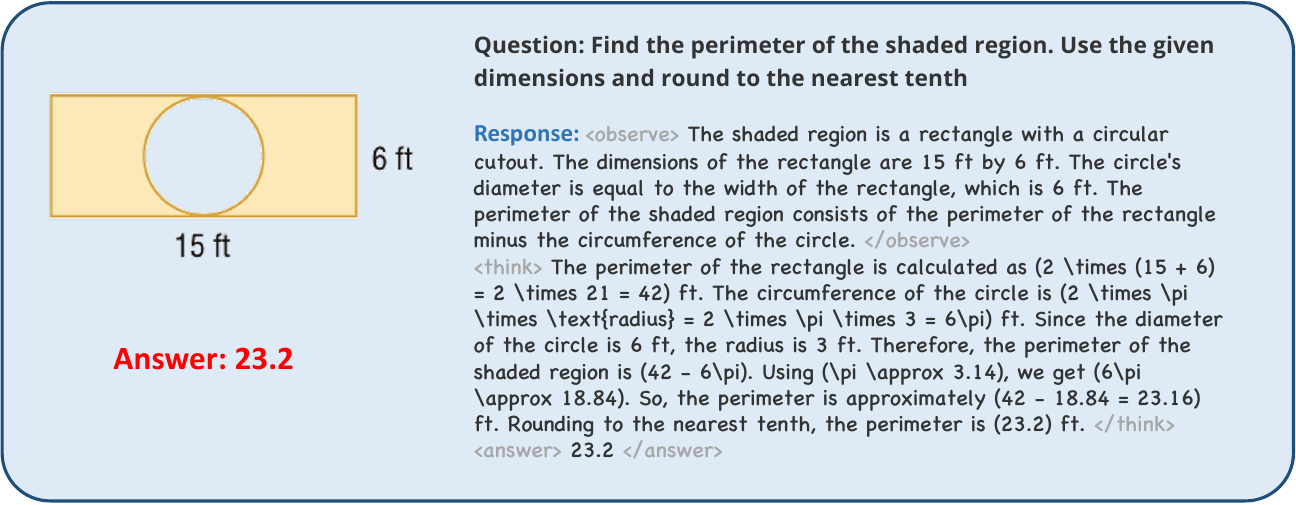}}
  \caption{A response example from Observe-R1.}
  \label{example4}
  \end{center}
  \vskip -0.15in
\end{figure*}

\begin{figure*}
  \begin{center}
  \centerline{\includegraphics[width=\columnwidth]{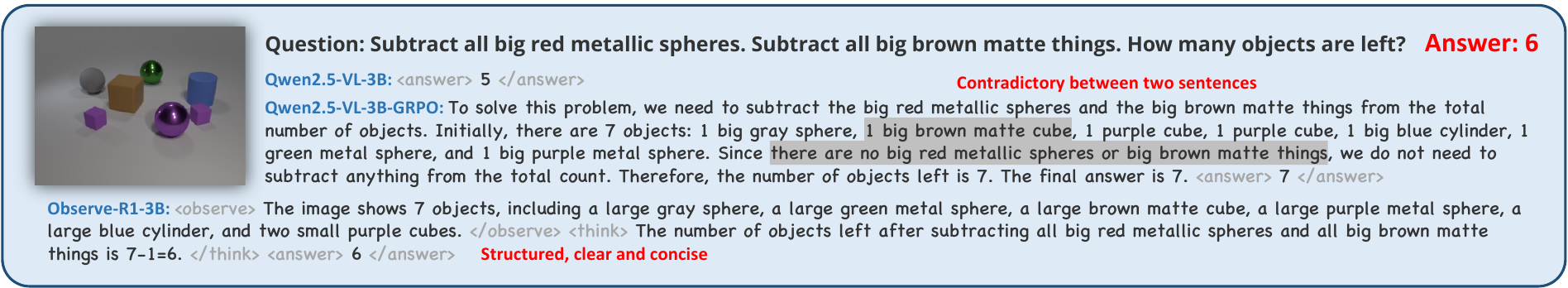}}
  \caption{An example from MathVista. Text with a grey background highlights the contradictions between the two sentences. The GRPO baseline accurately observes the image but fails to retain its content during reasoning.}
  \label{example1}
  \end{center}
  \vskip -0.15in
\end{figure*}

We conduct our experiments on Qwen2.5-VL-3B-Instruct and Qwen2.5-VL-7B-Instruct, randomly sampling 20k instances from the NeuraLadder dataset for RL training. We use 8*A100 80G GPUs for our model training. We set the rollout batch size to 128/128 and the training batch size to 64/128 for 3B/7B, respectively, with each sample generating eight responses. The learning rate is set to 1e-6/5e-7 for 3B/7B, respectively, the reward coefficient $\gamma_1$ and $\gamma_2$ are set to 0.5 and 0.2, respectively and $\sigma$ in the dynamic weight is set to 1.8.

For model evaluation, we choose MathVista~\citep{lu2024mathvista}, MathVerse~\citep{zhang2024mathverse} and MMK12~\citep{meng2025mm} to demonstrate the reasoning abilities of our model and choose MMStar~\citep{chen2024we}, MMMU~\citep{yue2024mmmu} and MMBench~\citep{liu2024mmbench} to evaluate the general abilities of our model. Concretely, we use MMMU-val and MMBench-dev-en for MMMU and MMBench, respectively. Additionally, for MMMU and MMBench, we prompt the model to only provide the final answer without reasoning to evaluate the general abilities without reasoning.

\subsection{Main Results}
We present our comprehensive comparisons with recent SOTA MLLMs in Table~\ref{mainres}. The results indicate that our 3B model outperforms most larger models listed in the table, including closed-source models and recent 7-11B reasoning models, demonstrating the effectiveness of our strategies. Unlike the original GRPO, which sees performance degradation on general tasks such as MMMU and MMBench, our method maintains strong performance on these benchmarks. On the reasoning benchmarks, Observe-R1-3B achieves a score of 64.9 on MathVista, outperforming all open-source and reasoning models, despite having only 3B parameters. This validates the superiority of our method in enhancing the reasoning capabilities of MLLMs while preserving strong performance on general multimodal tasks. In Figure~\ref{example4}, we present an example generated by Observe-R1-7B, which demonstrates a clear observation of the image elements and reasoning process.

\begin{figure*}
  \begin{center}
  \centerline{\includegraphics[width=\columnwidth]{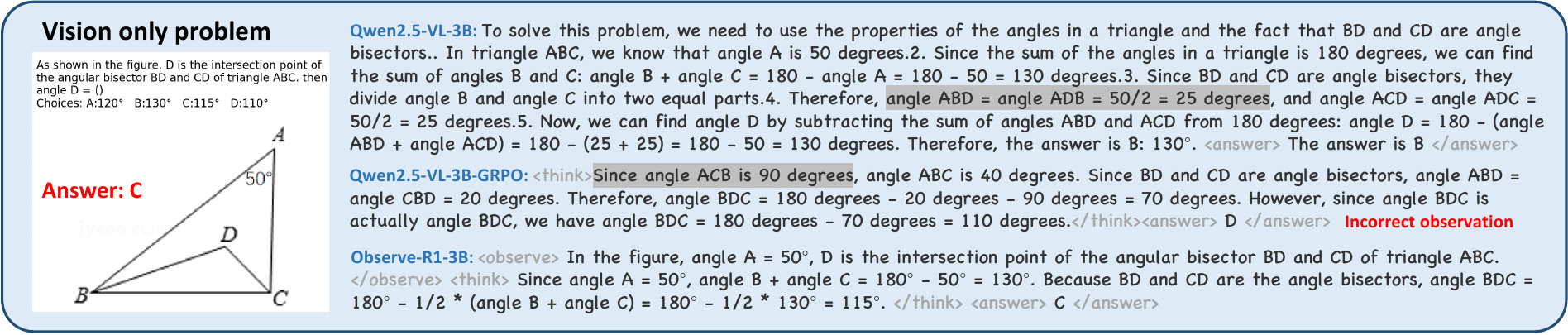}}
  \caption{A visual-only problem example from MathVerse. The GRPO baseline fails to observe the image accurately.}
  \label{example2}
  \end{center}
  \vskip -0.1in
\end{figure*}

In Figures~\ref{example1} and \ref{example2}, we present two examples from MathVista and MathVerse, showcasing the responses from three models for comparison. All three models are instructed to provide a reasoning process before providing the final answer. From Figure~\ref{example1}, we observe that the base model fails to follow the instructions and does not provide a reasoning process. In contrast, the RL fine-tuned model adheres to the instructions and accurately describes the image content. However, the GRPO model struggles to retain its content during the reasoning process due to an unstructured approach to multimodal data, while Observe-R1-3B presents a structured, clear, and concise reasoning process. In Figure~\ref{example2}, which presents a vision-only problem, we notice that the base model includes much repetitive information in its reasoning process. This redundancy hinders the model's ability to extract useful information clearly. The GRPO model fails to accurately describe the image, leading to an incorrect answer. In contrast, Observe-R1-3B can correctly describe the image content, benefiting from a multimodal format that encourages careful observation of the image.

\begin{table}[]
\caption{Ablation experiments of our proposed strategies. We report detailed results to fully validate the effectiveness of our method. For MathVista, we report five reasoning subtasks: SCI (scientific reasoning), ARI (arithmetic reasoning), GEO (geometric reasoning), ALG (algebraic reasoning) and STA (statistical reasoning). For MathVerse, results are categorized by modality emphasis: TD (Text Domain), TL (Text Lite), VI (Vision Intensive), VD (Vision Domain) and VO (Vision Only).}
\label{abres}
\centering
\resizebox{\columnwidth}{!}{%
\begin{tabular}{@{}l|cccccc|cccccc@{}}
\toprule
\multicolumn{1}{c}{\multirow{2}{*}{Model}} & \multicolumn{6}{c}{MathVista} & \multicolumn{6}{c}{MathVerse} \\ \cmidrule(l){2-13} 
\multicolumn{1}{c}{} & SCI & ARI & GEO & ALG & STA & ALL & TD & TL & VD & VI & VO & ALL \\\midrule
Baseline & 63.1 & 51.0 & 49.8 & 49.8 & 73.6 & 57.8 & 41.8 & 37.7 & 33.4 & 32.5 & 29.3 & 34.9 \\\midrule
Qwen2.5-VL-3B-GRPO & 60.7 & 56.4 & 57.3 & 57.3 & 75.4 & 60.0 & 48.2 & 39.7 & 35.7 & 34.3 & 33.1 & 38.2 \\
- with NeuraLadder & 62.3 & 58.1 & 60.3 & 60.9 & 78.4 & 63.2 & 46.1 & 41.0 & 38.1 & 34.8 & 34.6 & 38.9 \\
- with Mformat & 55.7 & 59.2 & 60.3 & 59.1 & 76.7 & 62.0 & 47.4 & 41.1 & 37.7 & 36.2 & 33.9 & 39.3 \\
- with Bonus & 59.8 & 57.9 & 59.4 & 59.4 & 75.8 & 61.4 & 47.8 & 40.7 & 36.1 & 34.6 & 33.3 & 38.5 \\
- with Dynamic & 61.5 & 60.0 & 58.4 & 57.7 & 77.1 & 62.7 & 46.2 & 41.4 & 38.0 & 35.7 & 33.4 & 38.9 \\\midrule
\rowcolor{gray!30} Observe-R1-3B & 63.7 & 58.2 & 65.3 & 63.2 & 77.6 & 64.9 & 48.5 & 42.1 & 39.7 & 37.8 & 35.4 & 40.7 \\ \bottomrule
\end{tabular}%
}
\end{table}

\subsection{Ablation Study}
In the ablation section, we conduct our experiments using Qwen2.5-VL-3B mainly on the reasoning tasks, exploring the effectiveness of our strategies in improving the reasoning capabilities of MLLMs. We aim to answer the following research questions.

\begin{figure}
  \begin{center}
  \subfigure[]{\includegraphics[width=.245\columnwidth]{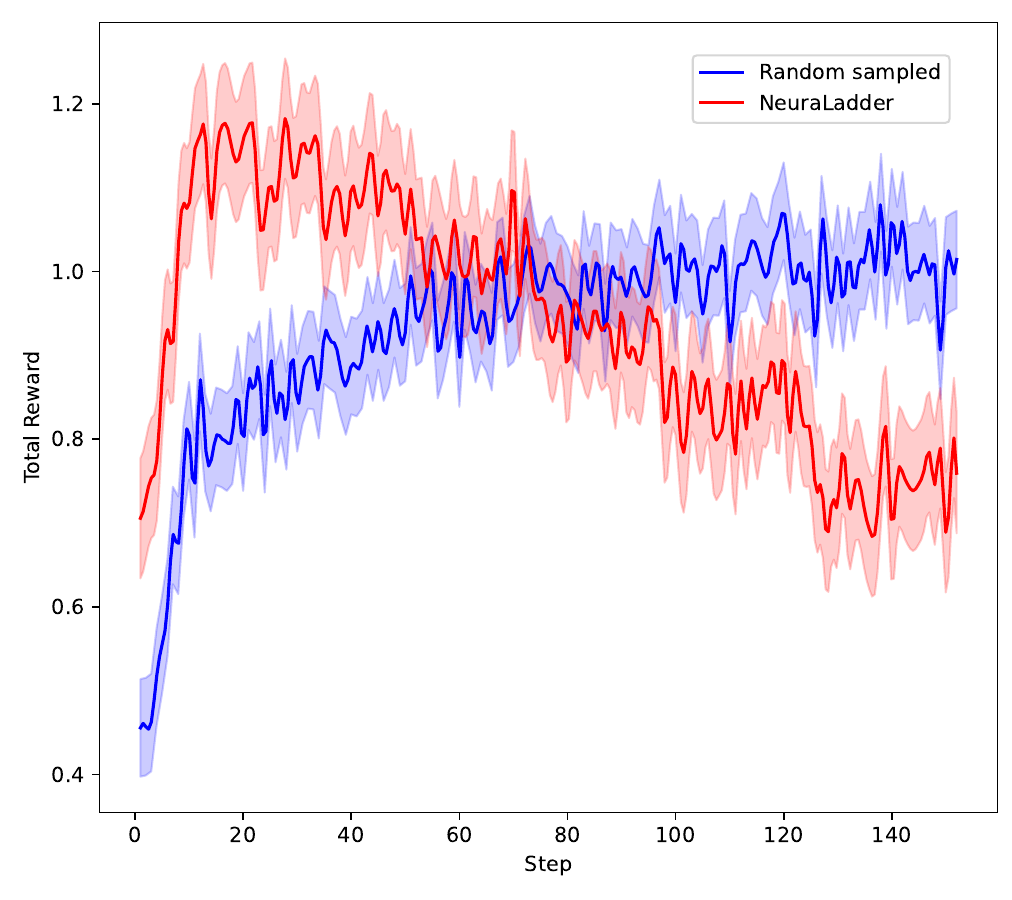}\label{rrspon:a}}
 \subfigure[]{\includegraphics[width=.245\columnwidth]{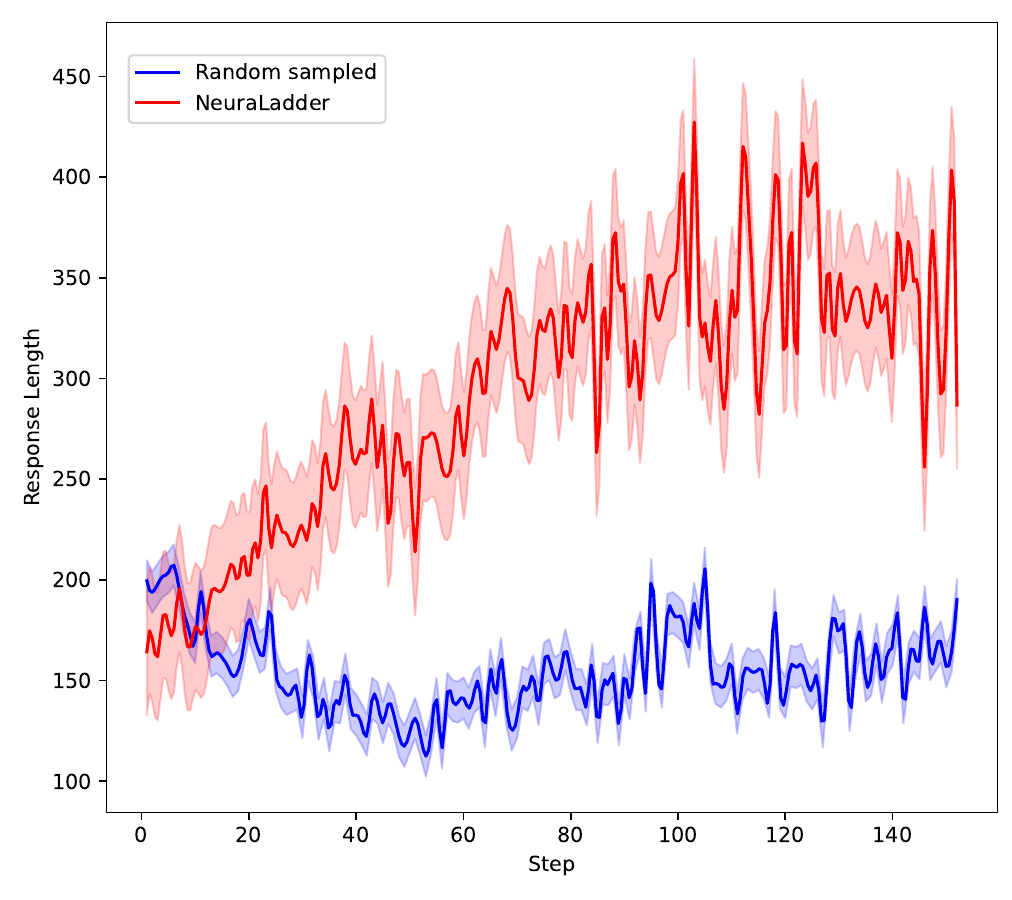}\label{rrspon:b}}
 \subfigure[]{\includegraphics[width=.245\columnwidth]{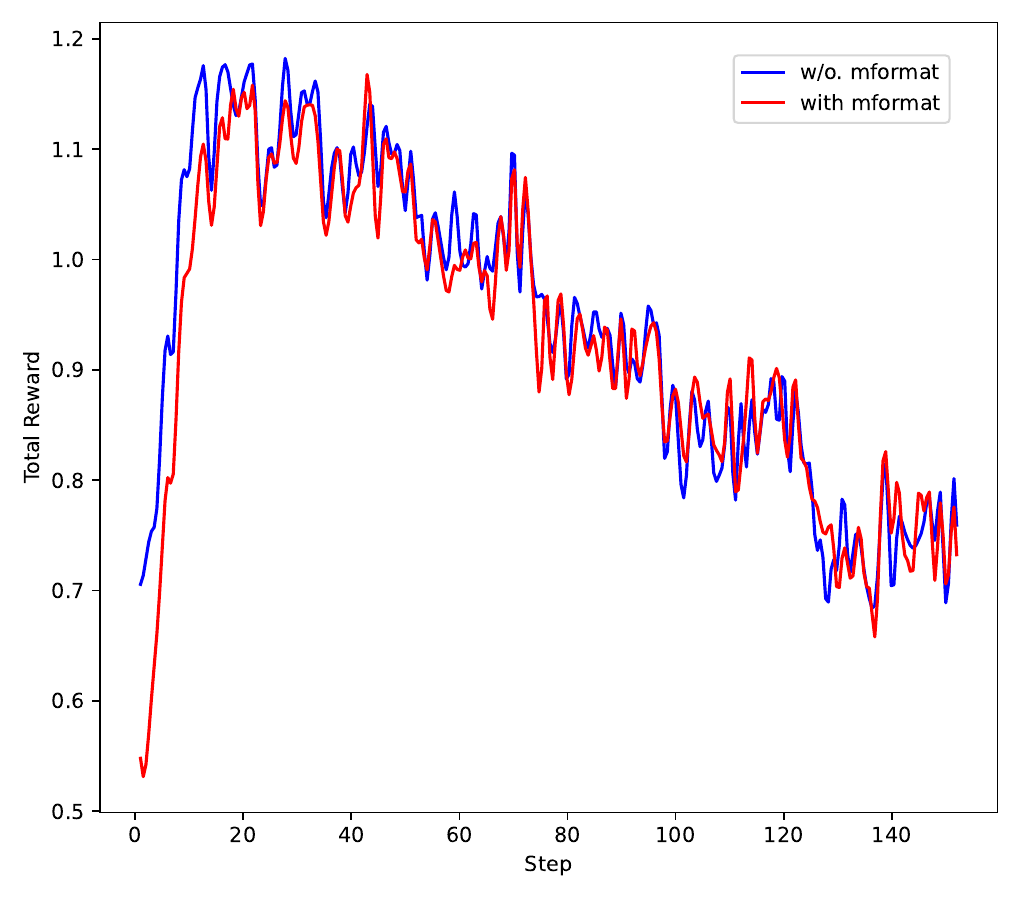}\label{rrspon:c}}
 \subfigure[]{\includegraphics[width=.245\columnwidth]{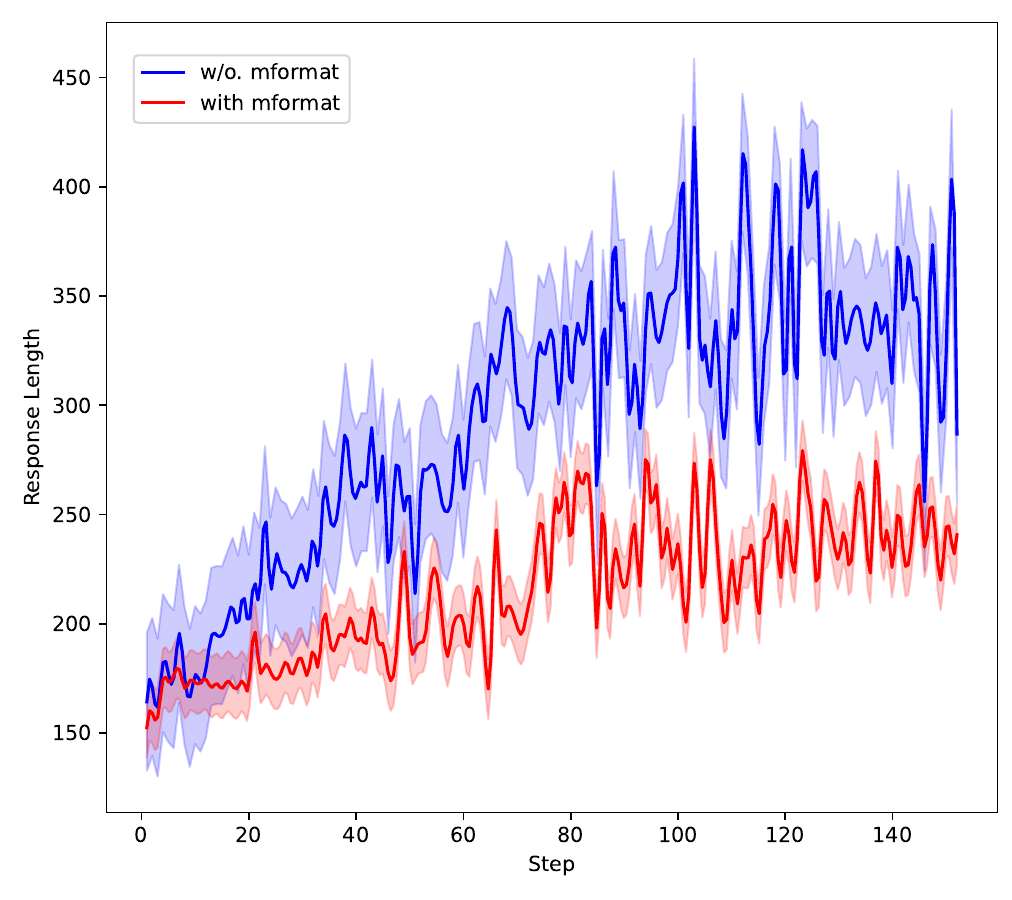}\label{rrspon:d}}
\\
  \caption{(a) Total rewards on NeuraLadder and randomly sampled dataset. (b)  Response length on NeuraLadder and randomly sampled dataset. (c) Total rewards with multimodal format. (d) Response length with multimodal format. We get these results using Qwen2.5-VL-3B on the same set of data. If the answer and format are both correct, the total reward will be 1.5.}
  \label{rrspon}
  \end{center}
\end{figure}

\textbf{RQ1: Why NeuraLadder and learn progressively?} One of the main contributions of our paper is the NeuraLadder dataset. We propose a progressive learning approach, enabling the model to reason step by step while also learning incrementally. As shown in Table~\ref{abres}, this progressive learning method results in a 3.2\% improvement on MathVista and a 0.7\% improvement on MathVerse compared to training on randomly sampled data. Additionally, we present the changes in total rewards and response lengths for both random and progressive learning in Figures~\ref{rrspon:a} and \ref{rrspon:b}. We observe that, when training on the NeuraLadder dataset, total rewards initially increase but later decrease, while response lengths continue to rise. In contrast, training on randomly sampled data results in steadily increasing total rewards, with response lengths fluctuating. The trends in the reward curves are easily understood, as the NeuraLadder dataset is organized by the difficulty of the samples. In the latter half of the curve, the model encounters a series of highly challenging data samples—questions generated by Qwen2.5-VL-7B that multiple attempts have failed to answer correctly. However, we can observe that, excluding the format reward of 0.5, the accuracy reward for the 3B model can still reach around 0.3 to 0.5. This demonstrates that, as training progresses, the model's reasoning ability continually improves, transforming originally difficult questions into moderately challenging ones. At the same time, even when faced with very difficult questions, the accuracy reward is still not significantly lower than that of the random sampling curve. From Figure~\ref{rrspon:b}, we can observe that the model trained on the NeuraLadder dataset, where response lengths continue to increase, finds it much easier to learn how to reason step by step compared to the model trained on the randomly sampled dataset. These findings validate the effectiveness of the NeuraLadder dataset and our approach to learn progressively.

\begin{figure*}
  \begin{center}
  \centerline{\includegraphics[width=\columnwidth]{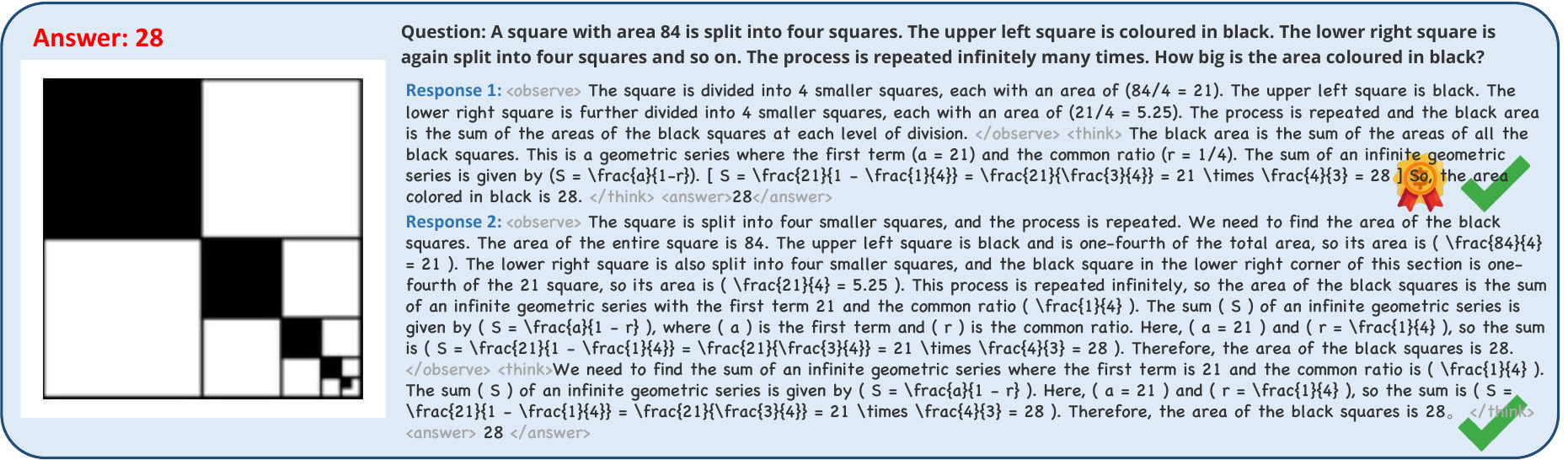}}
  \caption{A problem example from the training set where both answers are correct. Response 2 contains many repetitive contents and only obtains the accuracy and format rewards, while response 1 obtains an additional bonus reward for clearer and more concise reasoning process.}
  \label{example3}
  \end{center}
  \vskip -0.25in
\end{figure*}

\begin{wrapfigure}{r}{5cm}
\centering
\includegraphics[width=0.35\textwidth]{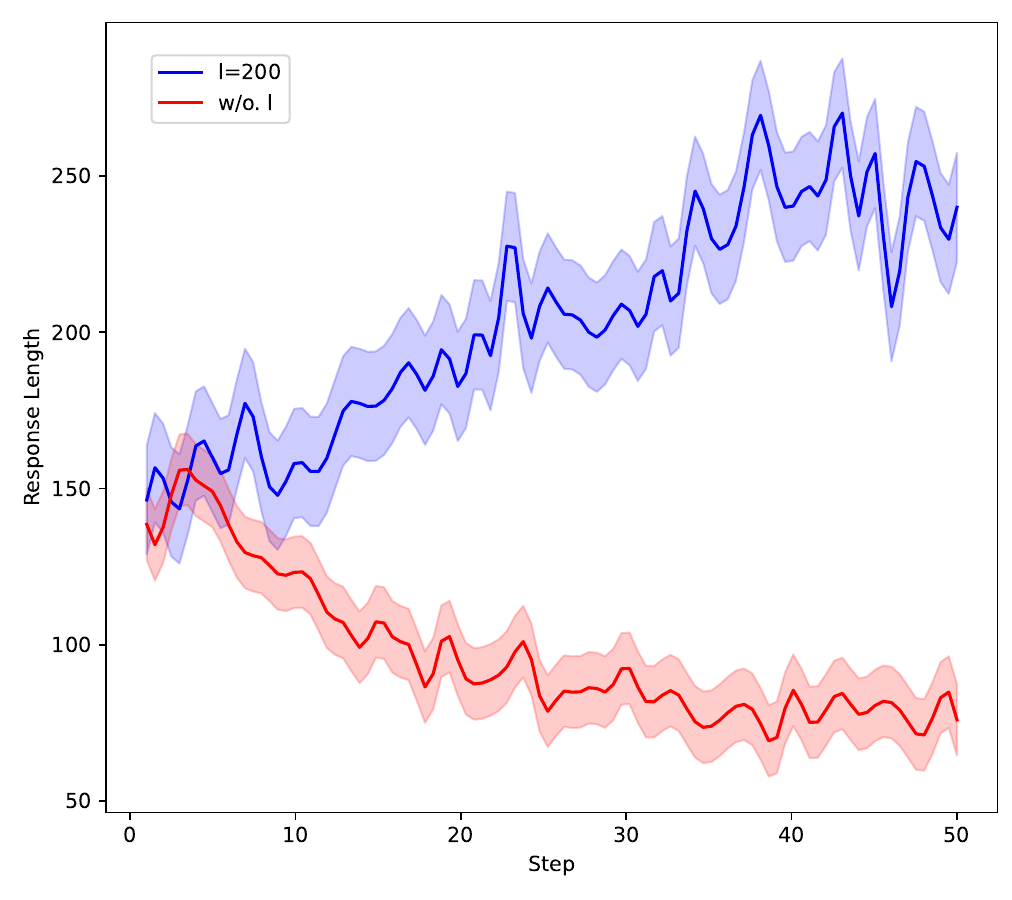}
\caption{Potential risk when using bonus reward.}
\label{abbonus}
\end{wrapfigure}

\textbf{RQ2: What role does the multimodal format play during the training process?} Firstly, the multimodal format makes the reasoning process more structured, concise, and clear. As shown in Figures~\ref{example1} and \ref{example2}, the model first extracts useful information from the image content based on the question, then reasons step by step, and finally provides the answer, avoiding redundant information that could hinder the reasoning process. Secondly, it encourages the model to observe visual information carefully through the \texttt{<observe></observe>} tag. In Figure~\ref{example2}, which presents a vision-only question, Observe-R1 accurately describes the image, while the GRPO model, lacking this format, fails to do so. Thirdly, the multimodal format leads to greater improvements, as shown in Table~\ref{abres}. Furthermore, from Figure~\ref{rrspon:c}, we observe that the total rewards are initially lower than those of GRPO but become slightly higher later. This is because the multimodal format is more difficult to learn than the GRPO format, resulting in lower rewards at first. As training progresses, both models adapt to the format, with the multimodal model achieving higher accuracy rewards and, consequently, higher total rewards.
Finally, the multimodal format enhances training efficiency. We find that training a model with this format is significantly faster than training without it, as the model typically generates clearer and more concise responses as illustrated in Figure~\ref{rrspon:d}. Therefore, it leads to a shorter training time.

\textbf{RQ3: The effectiveness of the plug-and-play bonus reward and potential risks.} The bonus reward is not designed to improve performance, but rather to encourage a better reasoning process. As shown in Figure~\ref{example3}, both response 1 and response 2 are correct; however, only response 1 is awarded the bonus reward. Response 2 contains much redundant information and many repetitive elements, while response 1 demonstrates a clearer and more concise reasoning process, which is encouraged for correct responses. However, there is a potential risk associated with using the bonus reward, as illustrated in Figure~\ref{abbonus}. When we discard the length constraint $\ell$ in Equation~\ref{eq5}, the response length might decrease. This occurs because when the model fails to follow the instructions for the reasoning process and only provides the correct final answer, it still receives the bonus reward, leading it to favor such responses. This risk can be effectively mitigated by adjusting the length constraint $\ell$. When $\ell$ approaches infinity, the bonus reward strategy reverts to the original strategy. A high $\ell$ means bonus rewards are only given for questions with higher average complexity, excluding lower complexity questions, even if answered correctly. We can adjust $\ell$ to meet different objectives.

\textbf{RQ4: The effectiveness of the dynamic weighting and sampling mechanism.} As shown in Table~\ref{abres}, the dynamic weighting and sampling strategy provides the second-best improvement among all strategies, highlighting its effectiveness. This approach employs a function that not only adjusts the types of data samples for more effective learning and exploration of reasoning but also dynamically filters samples to enhance training stability. As illustrated in Figure~\ref{rrspon:a}, we observe that during the training stage, original hard problems gradually transition to medium difficulty, allowing the dynamic weighting to focus more on these samples, which indicates a higher utilization of the data.

\section{Conclusion}
In this paper, we propose Observe-R1, a novel RL framework aimed at enhancing the reasoning capabilities of MLLMs. We introduce several key innovations, including progressive learning, a multimodal format, bonus rewards, and dynamic weighting and sampling. Additionally, we construct the NeuraLadder dataset based on the difficulty and complexity of the questions. Observe-R1 consistently outperforms a range of closed-source, open-source, and reasoning MLLMs on both reasoning and general multimodal tasks, demonstrating the effectiveness of our approach.

\textbf{Limitations and Future Directions. }
Due to limited computational resources, we are unable to conduct detailed experiments to further explore our strategies, including the use of larger MLLMs, more dynamic weighting functions, and the impact of bonus reward weights. 

Based on our current work, future directions include further investigation into the bonus reward and developing a more structured format for multimodal data. For example, it is promising to determine what types of correct answers and reasoning should receive additional bonus rewards. We believe that future research will delve deeper into these areas.

{
\small

\bibliography{ref}
\bibliographystyle{plainnat}
}

\end{document}